# HIERARCHICAL EVIDENCE ACCUMULATION IN THE PSEIKI SYSTEM and EXPERIMENTS IN MODEL-DRIVEN MOBILE ROBOT NAVIGATION[*]

A. C. Kak, K. M. Andress, C. Lopez-Abadia, M. S. Carroll and J. R. Lewis
Robot Vision Lab
School of Electrical Engineering
Purdue University
W. Lafayette, IN 47907

## ABSTRACT

In this paper, we will review the process of evidence accumulation in the PSEIKI system for expectation-driven interpretation of images of 3-D scenes. Expectations are presented to PSEIKI as a geometrical hierarchy of abstractions. PSEIKI's job is then to construct abstraction hierarchies in the perceived image taking cues from the abstraction hierarchies in the expectations. The Dempster-Shafer formalism is used for associating belief values with the different possible labels for the constructed abstractions in the perceived image. This system has been used successfully for autonomous navigation of a mobile robot in indoor environments.

## 1. APPLICATION

The PSEIKI system has been used successfully for autonomous navigation of a mobile robot in indoor environments. The series of experiments we ran this summer involved a spatial layout shown schematically from two different vantage points in Fig. 1. This figure is a depiction of the hallways in the lab area of our building, with doors, bulletin boards, etc., at various locations along the walls. The floor is made of semi-glossy tiles; these are a source of glare in camera images.

The mobile robot was provided with a 3-D model of the hallways using a CSG (constructive solid geometry) based representation. (The CSG approach to solid modeling uses a small number of volumetric primitives and regularized set-theoretic operators between them for constructing models of complex objects and shapes [Mortenson 1985].) If the mobile robot knows its position and orientation exactly, it can use ray casting to render an image corresponding to what the camera mounted on the robot should see. If, due to odometry errors, the position and the orientation of the robot are not known exactly, there will be a discrepancy between what the camera should be seeing and what is actually seen. It is important to realize that owing to the three dimensional geometry involved, this discrepancy will not be a simple translation of the expectation map with respect to the perceived image.

In these indoor navigation experiments, the mobile robot used PSEIKI for self-location. The robot would traverse a certain distance along one of the hallways, its uncertainty about its position and orientation increasing with travel, stop at the end of the distance, compare via PSEIKI the perceived image from that location with the expected image, and use trigonometry on the most believed abstractions derived from the perceived image to determine its location. Because of the computational costs involved, such exercises in self-location cannot be carried out continuously. How far the robot can go before it must self-locate is a function of the quality of the odometry and the maximum mis-registration that PSEIKI can tolerate for the purpose of "matching" the perceived image with the expectation map. In the experiments we have carried out so far, this distance is about 6 meters.

In many of our experiments, the robot was placed initially at location A in Fig. 2 and was asked to navigate autonomously to location B, using the supplied model of the hallways to figure out its initial path and to carry out self-locations whenever the position/orientation uncertainties

---

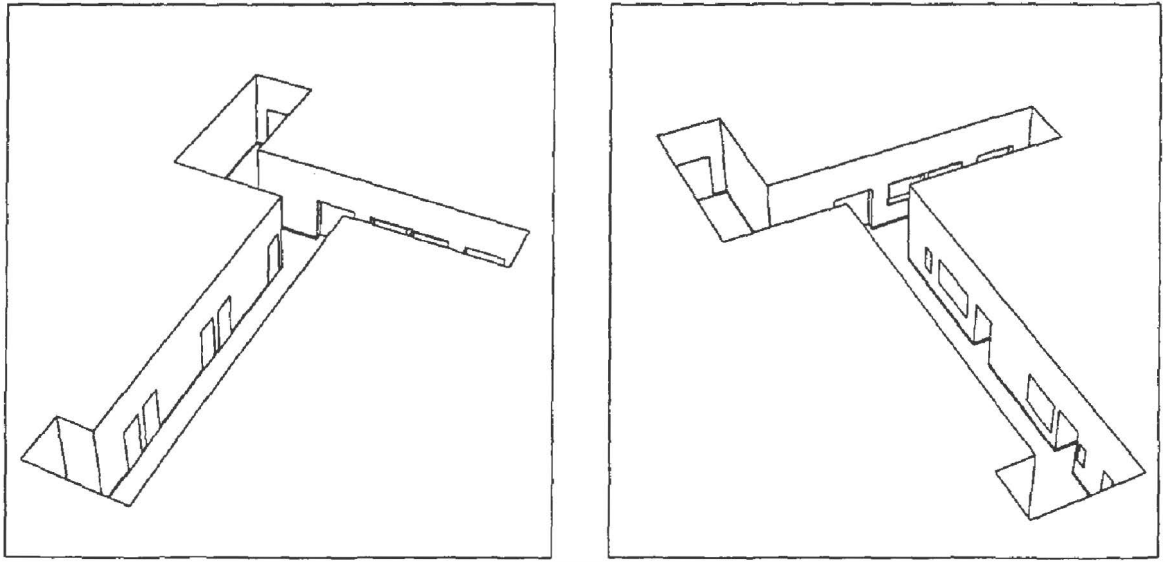

Fig. 1: Two different perspectives on the hallways used for vision-guided model-driven navigation experiments.

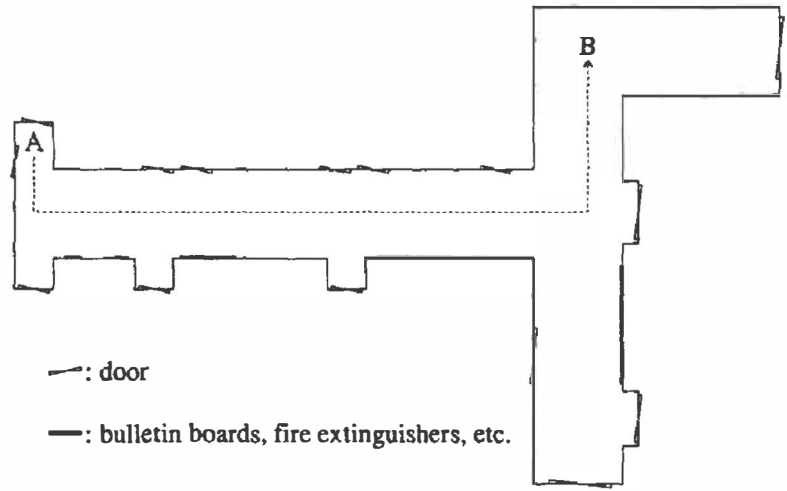

—: door

—: bulletin boards, fire extinguishers, etc.

Fig. 2: A plan view of the hallways.

exceeded a threshold. Of course, from the standpoint of path planning, the problem was trivial. The problem would also be trivial from the standpoint of navigating along the chosen path if the odometry were perfect, since then one could use the equivalent of "inertial" navigation and go in the blind from A to B. To cope with the uncertainties in odometry, the robot must self-locate every so often. To give the reader an idea of the quality of odometry on our robot, a commanded turn of 45° introduced in many instances an orientation uncertainty of 2° and a commanded straight-line motion had associated with it about 10% uncertainty in the location of the robot at the end of the motion. What's worse, due to uneven weight distribution in the base of the robot where a heavy battery is housed, a command to travel straight in a certain direction usually resulted in motion along a line that could be up to 15° off from the commanded direction. It was not possible for us to construct a usable model of this uncertainty as the uncertainties depended strongly on factors such as the starting orientation of the robot, whether or not the floor had been waxed recently, etc. We are

195

convinced that it would be impossible for our robot to navigate in the blind, without bumping into walls, from point A to point B shown in Fig. 2. The total distance between those two points is approximately 40 meters. To cope with the uncertainties introduced by poor odometry, we conservatively chose 6 meters as the longest distance the robot was allowed to travel without updating its location through PSEIKI.

In the rest of this paper, we first describe in the next section the geometrical representation used and the flow of control in PSEIKI. Next, in Section 3, we discuss how initial belief values are computed for the abstractions constructed in the perceived image and how these belief values are subsequently updated on the basis of relational considerations. Then, in Section 4, we comment on the issue of independence that is a prerequisite to the invocation of Dempster's rule for combining different belief functions. Subsequently, in Section 5, we briefly discuss how the output of PSEIKI is used for robot self-location. Finally, in our concluding remarks in Section 6, we present our opinion on whether it would be possible to use a relaxation based approach to accomplish what PSEIKI does.

The first detailed account on the evidence accumulation process and the flow of control in PSEIKI was published in [Andress and Kak 1988] and later updated in [Andress and Kak 1989]. The latter reference also includes more information on the CSG representation of 3-D models and how, for a given position and orientation of the robot, expectation maps are extracted from the CSG representation. Vision-guided model-driven experiments in mobile robot navigation using PSEIKI were first shown at the workshop cited in [Kak, Andress and Lopez-Abadia 1989]. Camera calibration and procedures for converting the output of PSEIKI into the position and orientation of the robot are discussed in [Lopez-Abadia and Kak 1989]. Finally, discussed in [Blask 1989] is a graphics tool that has been developed for debugging PSEIKI and testing its robustness.

## 2. REPRESENTATION AND FLOW OF CONTROL

PSEIKI has been implemented in OPS83 as a 2-panel / 5-level blackboard, as shown in Fig. 3. The left panel, called the *model panel*, holds the supplied abstraction hierarchy for the expectation map. A symbolic representation of the perceived image is deposited at the lowest levels of the right panel. In edge-based processing, the perceived image resides at the two lowest levels of the right panel, and, in region-based processing, there is also initial information at the face level. A uniform symbolic representation is used for all items that reside on the blackboard, regardless of panel or level. This symbolic representation consists of a data record containing many fields, one each for an *identity* tag, the name of the *panel*, the name of the *level*, etc. There is also a field that contains the identities of the *children* of the element, and a *value* field where information is stored on edge strength, average gray level in a region, etc. There also exists a set of fields for storing information on the parameters used for evidence accumulation.

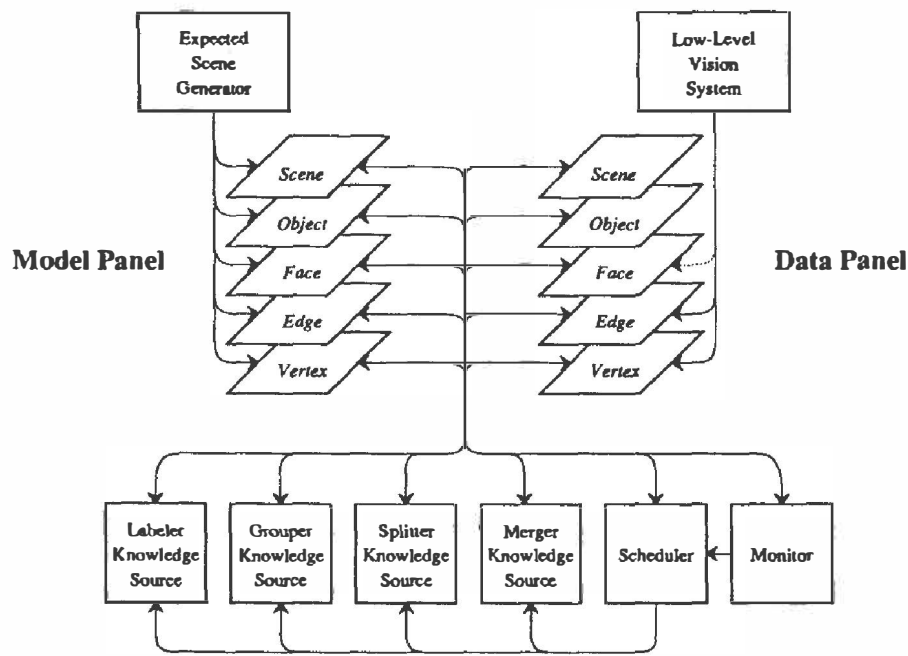

Fig. 3: The architecture of PSEIKI.

PSEIKI has four main knowledge sources (KS) that it uses to establish correspondences between the elements on the model side and the elements on the data side: *Labeler, Grouper, Splitter,* and *Merger*. The Grouper KS determines which data elements at a given level of the hierarchy should be grouped together to form a data element at a higher level of abstraction. Grouping proceeds in a data-driven fashion in response to goals that call for the establishment of nodes corresponding to the nodes on the model side. To explain, assume that the information shown in Fig. 4 resides on the blackboard. In this case, the nodes $F_A$, $F_B$ and $F_C$ will reside at the face level of the model panel. The node $F_A$ will point to the nodes $E_A$, $E_B$, $E_C$ and $E_D$ at the edge level, and so on. Of course, there will also be, on the model panel, a node at the object level pointing to the nodes $F_A$, $F_B$ and $F_C$ at the face level. In this case, the Scheduler, to be discussed later, posts goals that seek to establish data-panel nodes corresponding to the model nodes. For example, a goal will be posted to form data nodes, each node being a different hypothesis, for the model node $F_A$. To respond to this goal, the Scheduler will examine all the knowledge source activation records (KSAR's) that try to invoke the Grouper KS for those orphan data elements whose current labels correspond to one of the edges in $F_A$. (As soon as the Monitor sees a data element without a parent, it sets up a KSAR that seeks to invoke the Grouper KS.) The Scheduler will select that KSAR whose data edge has the strongest attachment with any of the edges in $F_A$ on the basis of the belief values. The data edge corresponding to such a KSAR then becomes a seed element for forming a grouping. In other words, the Scheduler uses this KSAR to fire the Grouper KS, which 'grows' the seed into an aggregation of data elements on the basis of relational considerations. For example, the Grouper KS will group $E_1$ with $E_3$ because the geometrical relationship between $E_1$ and $E_3$ is believed to be the same as between their current model labels. Using such considerations, for the example shown in the figure, the Grouper KS will propose the grouping $\{E_1, E_2, E_3, E_5, E_7, E_6, E_4\}$, under the face node $F_1$, and consider $F_1$ as a tentative correspondent of the model node $F_A$. This grouping will subsequently be examined for internal consistency by the Labeler KS for the purpose of computing our revised belief in each of the labels for the data edges in the grouping and in using $F_A$ as a label for $F_1$.

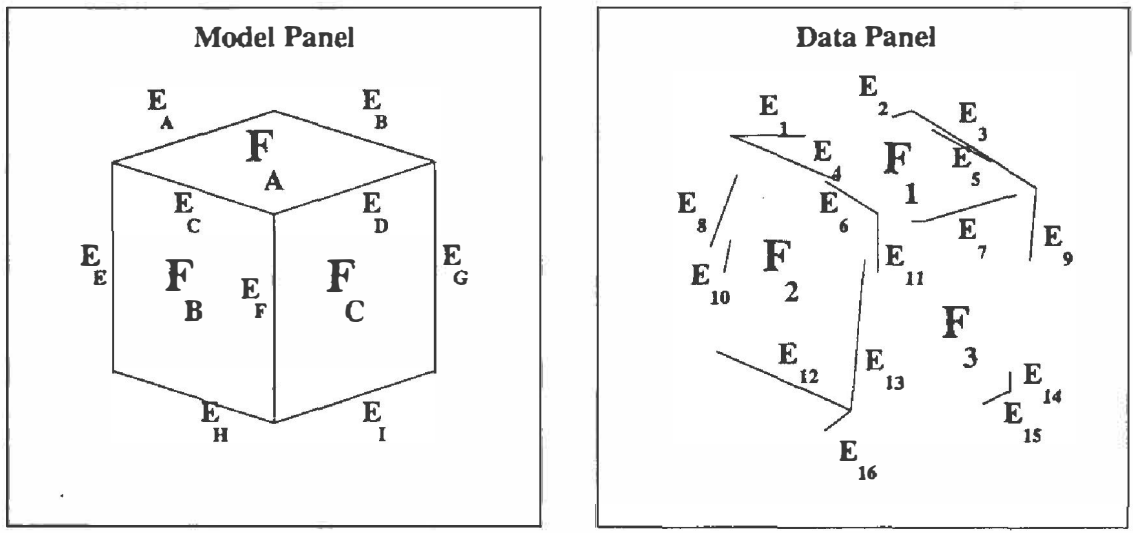

Fig. 4: We have used this example to explain many of the points regarding evidence accumulation in PSEIKI. The left frame depicts the information residing in the model panel, and the right frame the information in the data panel.

The first action by the Labeler KS, which takes place before the Grouper KS does any groupings at all, is to construct an initial frame of discernment (FOD) for each data element on the basis of physical proximity; meaning that initially all the model elements within a certain distance of the data element (at the same level of abstraction) will be placed in the FOD for the data element. Note that since the camera used is calibrated, the comparison between the model and the data takes place in the same space -- we could call it the image space, as opposed to the 3D space of the scene. The second major action of the Labeler KS, which takes place on a recurring basis, can be described as follows: Given a grouping from the model side and a tentative grouping on the data side, as supplied by the Grouper KS, the job of the Labeler KS is to estimate the degree of belief that can be placed in various possible associations between the data elements and their labels from the model side. These belief values are computed by revising the initial beliefs on the basis of the extent to which the data elements satisfy the relational constraints generated by their currently most-believed model labels. The revised beliefs are then propagated up the hierarchy, as discussed in the next section.

While the Grouper KS aggregates data elements at one level of abstraction for representation by a node at the next higher level, the function of the Merger KS is to aggregate data elements at one level of abstraction so that the aggregation can be treated as a single element at the *same* level. In other words, while the Grouper KS may group together a set of edges into a face, the Merger KS will try to group a series of short edges into a longer edge. The Splitter KS performs the opposite action of the Merger KS; it splits a single element on the blackboard into multiple elements at the same level.

The overall flow of control is controlled by the Monitor and the Scheduler, acting in concert. The Monitor uses OPS demons to run in the background, its task being to watch out for the data conditions that are needed for triggering the various KS's. For example, if there is a data element without a parent, it is the Monitor's job to become aware of that fact and synthesize a KSAR that is a record of the identity of the data element and the KS which can be triggered by that element. Initially, when the KSAR's are first created, they are marked as pending. When no KS is active, the Scheduler examines all the pending KSAR's and selects one according to prespecified policies. For example, the status of a KSAR that tries to invoke the Merger or the Splitter KS is immediately changed to active. It seems intuitively reasonable to fire these KS's first because they seek to correct any misformed groups.

More precisely, the operation of the Scheduler can be broken into three phases. In the first phase, the *initialization* phase, which uses extensive backchaining, the Scheduler operates in a completely model-driven fashion for the establishment of nodes on the data side corresponding to the supplied nodes on the model side. If the Scheduler cannot find data correspondents of the model nodes, it posts goals for their creation. In other words, the Scheduler examines the model panel from top to bottom, checks whether there exist a certain pre-specified number, $n_G$, of data correspondents of each model node. If the number of data nodes corresponding to a model node is fewer than $n_G$, the Scheduler posts goals for the deficit.

If this model-driven search is being carried out at a level that is populated with data nodes, then the Scheduler must initiate action to search through those data nodes for possible correspondents for the model node. This is done by activating the KSAR's that seek to invoke the Labeler KS for computing the initial belief values for the data elements, using only proximity considerations as discussed in the next section, and retaining up to $n_G$ data nodes that acquire the largest probability mass with respect to the model node. (Note that when the data elements are first deposited on the right panel of the blackboard, KSAR's for invoking the Labeler and the Grouper KS's for these data elements are automatically created; these KSAR's have pending status at the time of their creation.) For example, if the contents of the two panels of the blackboard are as shown in Fig. 4, the Scheduler will backchain downwards through the model panel, starting with the scene node. At the edge level, it will discover data on the right panel. The Scheduler will therefore activate the KSAR's that seek to compute the initial belief values for these edges. After the initial belief values are computed, the Scheduler will retain $n_G$ data edges for each model edge.

As was mentioned before, if the number of data nodes corresponding to a model node is fewer than $n_G$, the Scheduler posts a goal for the deficit. For example, for the case of Fig. 4, the Scheduler will recognize that initially there will not be any data nodes corresponding to the object level model node for the cube, so the Scheduler will post a goal for the establishment of $n_G$ object level data nodes for the cube. These $n_G$ nodes, after they are instantiated, will presumably lead to different and competing hypotheses (different groupings) at the object level. In the same vein, the Scheduler will post a goal for the establishment of $n_G$ competing nodes that would correspond to the node $F_A$. Since for the example under consideration, there exist data nodes at the edge level, the goals set up by the Scheduler would be somewhat different. For example, the data edges $E_4$ and $E_6$ may have edge $E_C$ as their labels. Therefore, the goal posted by the Scheduler will only require the establishment of $n_G-2$ additional data nodes corresponding to the model edge $E_C$. Of course, if no additional data edges can be found that can take the label $E_C$, the Scheduler will make do with just 2. This process is akin to using a depth bound for finding a solution in a search graph.

It should be clear that in the initialization phase, the operation of the Scheduler combines top-down model-driven search for grouping and labeling with bottom-up data-driven requests for finding parents for ungrouped data elements and for computing beliefs in the possible labels for the new groupings. Combinatorial explosions are controlled by putting an upper bound on the number of competing hypotheses that can be entertained in the model-driven search. It is important to note that the number of competing hypotheses for any model node is not limited to $n_G$. To explain, assume that the Grouper KS has grouped the edges $\{E_1,E_2,E_3,E_5,E_7,E_6,E_4\}$. Since the Splitter KS is given a high priority by the Scheduler, most likely this KS will fire next and probably discover that in the group formed data edge $E_3$ is competing with the data edge $E_5$. Therefore, the Splitter KS will split the group into two groups $\{E_1,E_2,E_5,E_7,E_6,E_4\}$ and $\{E_1,E_2,E_3,E_7,E_6,E_4\}$. In other words, because of the action of the Splitter KS, there can be a geometrical multiplication of the hypotheses formed by the Grouper KS. For these reasons, it becomes necessary to give a small value to $n_G$. For most of our experiments, $n_G$ is set to 3.

Since our explanation above was based on Fig. 4, the reader is probably wondering about how the Grouper KS might construct $n_G$ different and competing data groupings corresponding to, say, the model node $F_A$. After the first grouping is constructed by the procedure already discussed, the Scheduler will discover that it still does not have $n_G$ groupings corresponding to the model node $F_A$. As before, the Scheduler will examine all the pending KSAR's that seek to invoke the Grouper KS on data elements whose labels come from the edges in $F_A$. Of these, the KSAR associated with the data edge that attaches most strongly, on the basis of the belief values, with one of the edges in $F_A$, is selected for firing the Grouper KS, the data edge serving as a seed. (Note that the KSAR selected for the second grouping will not be the same as for the first grouping, since the KSAR used earlier is no longer pending.) After the second grouping is formed, it is compared with the first. If the two are identical, it is discarded. This process is continued until as many groupings can be formed as possible, with the total number not exceeding $n_G-1$ at the last attempt. When the $n_G$th grouping is formed, it is possible that owing to the action of the Splitter KS we may end with more than $n_G$ groupings.

Any time a new group is formed, a KSAR is created that seeks to invoke the Labeler KS for computing the initial belief values to be assigned to the data node corresponding to this grouping. For example, suppose the Grouper KS has formed the grouping $\{E_6,E_8,E_{10},E_{12},E_{13},E_{11}\}$ under the face node $F_2$ to correspond to the model face node $F_B$. Subsequently, the Labeler KS will construct a frame of discernment for $F_2$, consisting of all the model faces that have any overlap with $F_2$. In our example, this frame of discernment for $F_2$ could be $\{F_A,F_B,F_C\}$. The label assigned to $F_2$ will then consist of that model face label which gets the most mass using the formulas shown in the next section. It might seem incongruous to the reader that while the model face $F_B$ was used for constructing the grouping $F_2$, we should now permit the latter to acquire a different label. While in practice such a transfer of labels is not very likely, such a possibility has to be left open for the sake of a homogeneous computational procedure.

At the end of the initialization phase, the system has deposited on the data panel a number of competing nodes for each node on the model side. In practice, if the expectation map and the perceived image are sufficiently dissimilar, there will exist model nodes with no correspondents on the data side. At the same time, especially if the image pre-processor is producing many parallel lines for each real edge in the scene, there will exist many competing nodes, possibly exceeding $n_G$, on the data side for each node on the model side. It is important to note that the labels generated for the data nodes in the initialization phase of the Scheduler only involved proximity consideration. Relational considerations are taken into account in the phase discussed next.

The second phase of the Scheduler is the *updating* phase. Unlike in the first phase, during the updating phase the Scheduler makes no use of the contents of the model panel. On the other hand, the Scheduler traverses the data panel from top to bottom and invokes relational considerations through the Labeler KS to revise the belief values in the association of the data nodes and their labels. Of course, the Labeler KS must access the model information to figure out the geometrical relationships between the different model nodes, so that these relationships can be compared with those between the corresponding data nodes. To explain, let's go back to the example of Fig. 4. During the initialization phase when the Labeler assigns initial beliefs to the nodes in the data panel, it also creates KSAR's for updating these belief values; however, these KSAR's are not attended to by the Scheduler until the updating phase. For the example of Fig. 4, the Scheduler will first look at the KSAR corresponding to the object level nodes in the data panel. Consider the object level node made of the face grouping $\{F_1,F_2,F_3\}$. The KSAR that calls for revising the beliefs associated with this object level node will in fact apply the Labeler KS to the face grouping $\{F_1,F_2,F_3\}$ using relational considerations such as similarity of the transformations between $F_1$ and $F_2$ on the one hand and $F_A$ and $F_B$ on the other, assuming that $F_A$ and $F_B$ are the current labels for $F_1$ and $F_2$, respectively. Similarly, when the KSAR for updating the belief values associated with the face $F_2$ is processed, the result is the application of the Labeler KS to the edges $\{E_6,E_8,E_{10},E_{12},E_{13},E_{11}\}$ for belief revision on the basis of relational considerations.

Finally, during the last phase, the *propagation* phase, the belief revision takes place by propagating the belief values up the data panel hierarchy.

Although the operation of the Scheduler was presented as consisting of three separate phases, temporally speaking the boundaries between the phases are not as tight as what might be construed by the reader from our discussion so far. For example, if in the middle of the updating phase the labels of two faces become identical and if these faces satisfy certain additional criteria, such as sharing a common boundary, the Merger KS will merge the two faces into a single grouping. When the Merger KS creates this new grouping, it will also post KSAR's for invoking the Labeler KS for initial belief value computations. This is one example of how computations typical of the initialization phase may have to be carried out during another phase. Another example would be when a data node changes its label during the process of belief revision in the update phase. Note that a data node takes that label for which it has the largest probability mass in the frame of discernment. The process of updating beliefs on the basis of relational considerations can lower the belief in the currently held label for a data node vis-a-vis the other labels in the frame of discernment. When that happens, the data node will change its label and that would trigger the formation of KSAR's of the updating and initialization variety. For example, for the case of Fig. 4, suppose during the update phase the label for the data edge $E_1$ changes from $E_A$ to $E_C$. This would trigger the formation of a KSAR for updating the belief in the new label $E_C$. Similarly, if $F_1$'s label were to change from $F_A$ to, say, $F_C$ during the update or the propagation phases, that would launch a KSAR that we refer to as the "labeling KSAR with re-labeling action." Assuming that at the instant $F_1$'s label changed, its children were $\{E_1,E_2,E_3,E_5,E_7,E_6,E_4\}$, the re-labeling action consists of first eliminating any previous bpa's (basic probability assignment functions) and frames of discernment for all of these $E_i$'s, and then using the edges in $F_C$ as the new frame of discernment for each $E_i$.

## 3. ACCUMULATION OF EVIDENCE

Evidence accumulation in PSEIKI is carried out by the Labeler KS, which invokes different procedures for each of the three phases of the Scheduler. As opposed to being formalistic, practi-cally all our explanation in this section will be with the help of simple examples. A more formal exposition can be found in [Andress and Kak 1989].

*Initialization:*

Recall that in the initialization phase, the Labeler KS is called upon to examine different possible associations between the data nodes and the model nodes, the model node candidates for such associations being determined solely on the basis of their physical proximity to the data nodes. Let's say that during the initialization phase, an initial bpa function is sought for the data edge $E_1$ in Fig. 4. The Labeler KS will pool together all the model edges whose centers of mass are within a radius $r_{max}$ of the center of mass of $E_1$ and call this pool the frame of discernment for figuring out the labels for $E_1$. Let's say that this FOD, denoted by $\Theta_{initial}$, consists of

$$\Theta_{initial} = \{E_A, E_C, E_E\} \quad (1)$$

To accumulate belief over this FOD, we use the metaphor that each model edge in the FOD is an expert and tells us, by using similarity and dissimilarity metrics, how much belief it places in its similarity to the data edge $E_1$. In other words, the expert $E_A$ gives us the following information

$$m_{E_1}(\{E_A\}) = \text{similarity\_metric}(E_1, E_A)$$
$$m_{E_1}(\{\neg E_A\}) = \text{dissimilarity\_metric}(E_1, E_A)$$
$$m_{E_1}(\Theta) = 1 - m_{E_1}(\{E_A\}) - m_{E_1}(\{\neg E_A\})$$

(2)

As the reader may recall, the bpa shown constitutes what Barnett calls a *simple evidence function* [Barnett 1981]. The similarity metrics currently being used in PSEIKI are presented in [Andress and Kak 1989]. For the example under discussion, the "experts" $E_C$ and $E_E$ will yield the following two simple evidence functions:

$$m_{E_1}(\{E_C\}) = \text{similarity\_metric}(E_1, E_C)$$
$$m_{E_1}(\{\neg E_C\}) = \text{dissimilarity\_metric}(E_1, E_C)$$
$$m_{E_1}(\Theta) = 1 - m_{E_1}(\{E_C\}) - m_{E_1}(\{\neg E_C\})$$

(3)

and

$$m_{E_1}(\{E_E\}) = \text{similarity\_metric}(E_1, E_E)$$
$$m_{E_1}(\{\neg E_E\}) = \text{dissimilarity\_metric}(E_1, E_E)$$
$$m_{E_1}(\Theta) = 1 - m_{E_1}(\{E_E\}) - m_{E_1}(\{\neg E_E\})$$

(4)

The Labeler KS combines these simple evidence functions using Barnett's algorithm. The accumulated belief is computed only for the singleton propositions in $\Theta_{initial}$. The singleton proposition with the largest mass is then called the *current* label for the data edge $E_1$. Assume for the sake of discussion, that at this point the Labeler has declared $E_A$ to be the current label for $E_1$. *It is most important to note that the three bpa's shown above are not discarded at this point. During the update phase, when beliefs are being revised on the basis of relational considerations, the updating bpa's are combined with the bpa's shown above. Also, the FOD for the data nodes is expanded to include additional labels representing the model correspondents of those data groupings in which the data node is currently participating. After the updating bpa's are combined with the initial bpa's shown above, it is entirely possible that the largest probability mass will be accrued for a singleton proposition that is different from the currently held label for $E_1$. When this happens, the label for $E_1$ will automatically change to the one for which the probability mass is now maximum.*

In procedure, the computation of initial beliefs and labels at all levels of the blackboard is identical to that outlined above, only the similarity and dissimilarity metrics used are different.

*Belief Revision:*

To explain with an example the process of belief revision on the basis of relational considerations, let's assume that the Grouper has advanced $\{E_1, E_2, E_5, E_7, E_6, E_4\}$ as a possible grouping, under the face level node $F_1$, and that the current label for $F_1$ is $F_A$. (Operationally, the procedure for finding the *current* label for face $F_1$ is identical to the one described above under Initialization. The Labeler constructs an initial FOD for $F_1$ on the basis of physical proximity, uses face similarity metrics to generate a set of simple evidence functions for the singleton propositions in this FOD, and finally sets $F_1$'s label to the singleton proposition with the largest probability mass.) Let's now focus on the data edge $E_1$ from the grouping and explain what happens during the update phase of the Scheduler. First note that as soon as the $F_1$ grouping $\{E_1, E_2, E_5, E_7, E_6, E_4\}$ is formed, the FOD for edge $E_1$ is enlarged by adding to $E_1$'s initial FOD the model edge set corresponding to the face node $F_A$, since $F_A$ is the current label for $F_1$. This FOD enlargement is carried out for each $E_i \in F_1$. In other words, as soon as the grouping $F_1$ comes into existence, the following new FOD is formed for $E_1$:

$$\Theta_{revised} = \{E_A, E_C, E_E, E_B, E_D\} \qquad (5)$$

which is obtained by taking the union of the initial FOD for $E_1$ and the members of the grouping corresponding to the label $F_A$ for the face node $F_1$.

Therefore, when the Labeler is invoked with a KSAR seeking to update the belief value for a face node, such as node $F_1$, the Labeler understands the request to mean that beliefs should be revised for all the children of $F_1$ on the basis of their geometrical interrelationships vis-a-vis the corresponding relationships on the model side. The metaphor used for updating the beliefs associated with $E_1$ on the basis of its belonging to the grouping $F_1$ is that all the other edge elements in $F_1$ are "experts" in figuring out their geometrical relationships to the edge $E_1$ and comparing these relationships to those satisfied by their labels.

To elaborate, let's say that we want to compute the contribution that $E_5$ will make to revising our belief in the assertion that $E_1$'s label is $E_A$. To estimate this contribution, we will set up the following bpa:

$$m_{update: E_5 \to E_1}(\{E_A\}) = m_{E_5}(\{E_X\}) \cdot$$
$$\text{rel\_similarity\_metric}(E_5, E_1; E_X, E_A)$$

$$m_{update: E_5 \to E_1}(\{\neg E_A\}) = m_{E_5}(\{E_X\}) \cdot$$
$$\text{rel\_dissimilarity\_metric}(E_5, E_1; E_X, E_A)$$

$$m_{update: E_5 \to E_1}(\Theta_{revised}) = 1 - m_{update: E_5 \to E_1}(\{E_A\})$$
$$- m_{update: E_5 \to E_1}(\{\neg E_A\}) \qquad (6)$$

where $E_X$ is $E_5$'s current label; X could, for example, be B. The relational similarity and dissimilarity metrics, rel\_similarity\_metric and rel\_dissimlarity\_metric, give us measures of similarity and dissimilarity of the geometrical relationship between the first two arguments and the geometrical relationship between the last two arguments. For example,

$$\text{rel\_similarity\_metric}(E_5, E_1; E_X, E_A)$$

figures out the rigid body transformation between $E_5$ and $E_1$, figures out the rigid body transformation between $E_X$ and $E_A$, compares the two transformations, and then returns a measure of similarity between the two transformations. Further details on these rigid body transformations and their comparisons can be found in [Andress and Kak 1989].

For the example under consideration, for each index i in the set $\{2,5,7,6,4\}$, this updating process will generate the following simple evidence function:

$$m_{update: E_i \to E_1}(\{E_A\})$$
$$m_{update: E_i \to E_1}(\{\neg E_A\})$$
$$m_{update: E_i \to E_1}(\Theta_{revised}) \qquad (7)$$

When these simple evidence functions are combined using Dempster's rule with the simple evidence functions generated during the initialization phase, we obtain our revised belief in various possible labels for the data edge $E_1$. For the example under consideration, note that any belief in the proposition $\neg E_A$ will lend support to labels other than $E_A$ in the FOD of $E_1$. (Since the focal elements of all the update bpa's for, say, the data edge $E_1$ are the same, these in the current example being $\{E_A\}$, $\{\neg E_A\}$ and $\Theta_{revised}$, Dempster's rule for combining the bpa's possesses a simple and fast implementation without involving any set enumeration.) As was mentioned before, if during such belief revision the largest mass is accrued for a singleton proposition that is not the current label, then the current label would change and correspond to the singleton proposition.

The reader should note that the request to update the beliefs associated with the data node $F_1$ caused the beliefs and labels associated with $F_1$'s children to be altered. In other words, during the update phase, a KSAR ostensibly wanting to update the beliefs associated with the nodes at one level of abstraction actually causes the updating to occur at a lower level of abstraction. Although making for a cumbersome explanation of the belief revision process, there is an important operational reason for this. The belief revision process occurs on relational considerations, involving mutual relationships amongst the members of a data grouping vis-a-vis the corresponding relationships amongst the labels in a model grouping. Since the grouping information at one level of abstraction can only be determined by examining the nodes at the next higher level of abstraction, hence the reason for using the update KSAR's for, say, the face level nodes to actually update the beliefs associated with the nodes at the lower level, the edges.

Before leaving the subject of belief updating, we would like to mention very quickly that when an edge like $E_6$, which is common to two faces, is first grouped into, say, face $F_1$, its FOD is expanded by taking a union of its initial FOD and the edges that are grouped under the current label of $F_1$. If we assume that the initial FOD for $E_6$ was

$$\Theta_{initial} = \{E_C, E_D, E_F\}$$

and if we further assume that the current label for $F_1$ is $F_A$, then upon the formation of the first grouping, $E_6$'s FOD is revised to

$$\Theta_{revised} = \{E_F, E_A, E_B, E_D, E_C\}$$

Now, when $E_6$ is grouped again under the face node $F_2$, $E_6$'s FOD gets further revised to become

$$\Theta_{revised} = \{E_F, E_A, E_B, E_D, E_C, E_E, E_H\}$$

which is the union of the previous FOD and the group of model edges under $F_B$, assuming that $F_B$ is the current label for $F_2$. Since the bulk of grouping takes place during the initialization phase of the Scheduler, for such an edge the FOD used for belief revision using relational consideration would, in most cases correspond to the latter version.

*Propagation:*

During this phase, the Labeler "pushes" the belief values up the abstraction hierarchy residing in the data panel. The rationale on which we have based PSEIKI's belief propagation up the data hierarchy satisfies the intuitive argument that any evidence confirming a data element's label should also provide evidence that its parent's label is correct.

Continuing with the previous example, note that the request to update the beliefs associated with the face node $F_2$ actually caused the beliefs associated with $F_2$'s children to be revised on the basis of relational considerations. During the propagation phase, we want the revised beliefs associated with $F_2$'s children to say something about the beliefs associated with $F_2$ itself. To explain how we propagate the beliefs upwards, let's consider the nature of the bpa obtained by combining all the update bpa's for the data edge $E_1$:

$$m_{update:E_1} = m_{update:E_2 \to E_1} \oplus$$

$$m_{update:E_5 \to E_1} \oplus \ldots \ldots \quad (8)$$

If at the time of computation of the individual update bpa's here, the label for $E_1$ was $E_A$, then the focal elements of the function $m_{update:E_1}$ are only $\{E_A\}$, $\{\neg E_A\}$ and $\Theta_{revised}$. Clearly, the update probability mass as given by $m_{update:E_1}(\{E_A\})$ arises from the consistency of $E_1$'s label with its sibling's labels, all these labels being derived from the children of the face node $F_A$; this probability mass can therefore be considered as a weighted vote of confidence that face $F_1$'s label is $F_A$. Similarly, the update probability mass given by $m_{update:E_1}(\{\neg E_A\})$ arises from the inconsistency of $E_1$'s label with the labels of its siblings, the labels again being derived from the children of $F_A$; this mass can therefore be considered as a weighted vote of no confidence in the assertion that $F_1$'s label is $F_A$. In a similar vein, $m_{update:E_1}(\Theta_{revised})$ may be considered as a measure of ignorance about $F_1$'s current label, from the standpoint of the "expert" $E_1$, ignorance in light of the labels currently assigned to $E_1$ and its siblings. On the basis of this rationale, we can construct the following bpa for updating the beliefs associated with the face node $F_1$:

$$m_{update:E_1 \to F_1}(\{F_A\}) = m_{update:E_1}(\{E_A\})$$
$$m_{update:E_1 \to F_1}(\{\neg F_A\}) = m_{update:E_1}(\{\neg E_A\})$$
$$m_{update:E_1 \to F_1}(\Theta_{face}) = m_{update:E_1}(\Theta_{revised}) \quad (9)$$

Since $m_{update:E_1}$ is a valid bpa, having been obtained by the combination shown in Eq. (8), it follows that $m_{update:E_1 \to F_1}$ must also be a valid bpa. We can now express the total new accumulated belief for face $F_1$ from its children $E_1, E_2, E_5, E_7, E_6, E_4$ as

$$m_{update:F_1} = m_{update:E_1 \to F_1} \oplus$$

$$m_{update:E_2 \to F_1} \oplus m_{update:E_5 \to F_1} \oplus \ldots \ldots \quad (10)$$

The belief expressed by the bpa $m_{update:F_1}$ will be focussed on the propositions $\{F_A\}$, $\{\neg F_A\}$ and $\Theta_{face}$. This bpa is combined with the currently stored simple evidence functions for the face node $F_1$. Of course, if as a result of this combination, the probability mass assigned to the label $F_A$ is no longer the maximum, the label of $F_1$ will be changed, which, as was mentioned before, invokes the initialization type computations once again.

## 4. EDGE-BASED vs. REGION-BASED OPERATION

Note that PSEIKI can be operated in two different modes: the edge-based mode and the region-based mode. In the edge-based mode, edges extracted from the perceived image are input into the two lowest levels of the data panel of the blackboard in Fig. 3. In the region-based mode, the perceived image is segmented into regions of nearly constant gray levels and the result input into the three lowest levels of the data panel. Note that in the region-based mode, there is no presumption that the regions would correspond to the faces in the expectation map. In fact, in most imagery, because of glare from surfaces and other artifacts, each face in the expectation map will get broken into many regions in the data and there can also be regions in the data that can straddle two or more faces in the expectation map. However, as we have noticed in our experiments, in many cases the Merger KS is able to merge together some of the regions that correspond to the same face. Of course, it is not necessary for such merging to be perfect since for experiments in mobile robot self-location we do not need 1-1 correspondence between the perceived image and the expectation map everywhere. Thanks to the rigid-body constraints, the scene to model correspondence need only be established at a few locations to calculate the position and the orientation of the mobile robot with precision, as long as these locations satisfy certain geometrical constraints. More on this subject later.

## 5. ARE INDEPENDENCE CONDITIONS SATISFIED?

The following question is frequently raised regarding evidence accumulation in PSEIKI: Have we satisfied the necessary condition for the application of Dempster's rule, the condition that says that all evidence must come from disparate sources, i.e., the sources of evidence must be independent? Superficially it may seem that PSEIKI violates this condition since when we compute an update bpa, as for example shown in Eqs. (6), we multiply the relational metrics by an initial bpa.

Despite the aforementioned use of initial belief functions in the updating process, a closer examination reveals that the independence requirements are not being violated. To explain, let's first state that by independence we mean lack of predictability. Therefore, the question of independence reduces to whether an updating bpa, like $m_{update:E_5 \to E_1}$ in Eqs. (6), can be predicted to any extent from a knowledge of one of the initial bpa's, for example $m_{E_5}$. We believe such a prediction can not be made for two reasons: 1) A product of a predictable entity with an unpredictable entity is still unpredictable; and, 2) the relational metric values that enter the formation of $m_{update:E_5 \to E_1}$ are not within the purview of the "expert" giving us $m_{E_5}$ on the basis of non-relational and merely geometric similarity of a data node with those model nodes that are in spatial proximity to the data node.

Another way of saying the same thing would be that since PSEIKI may be called upon to "match" any image with the expectation map, it has no prior knowledge that the structure (meaning the relationships between different data entities) extracted in the supplied image bears any similarity to the structure in the expectation map. In general, it must be assumed that the data elements can be in any relationships -- and, therefore, unpredictable relationships -- vis a vis the relationships between their currently believed model correspondents. Therefore, it will not be possible to predict a probability mass distribution obtained from relational considerations from a probability mass distribution obtained from just element-to-element similarity considerations. Hence, we can claim that the independence requirements are not being violated for the application of Dempster's rule.

## 6. ROBOT SELF-LOCATION USING PSEIKI

Fig. 5a shows a line drawing representation of an expected scene that was used by PSEIKI for the interpretation of the image shown in Fig. 5b during a navigation experiment. The expectation map of Fig. 5a, obtained by rendering the CSG representation of the hallways using the calibration parameters of the camera on the robot and the position of the robot as supplied by odometry, was supplied to PSEIKI as a four level hierarchy of abstractions: vertices, edges, faces and scene. This abstraction hierarchy was produced by modifying a CSG based geometric modeler developed in the CADLAB at Purdue [Mashburn 1987]. Edges and regions were extracted from the camera image of Fig. 5b and input into the three lowest levels of the data panel of the blackboard.

When PSEIKI terminates its processing, at the end of the belief propagation phase, there will generally be multiple nodes at the scene level, each with a different degree of belief associated with it. The autonomous navigation module controlling the mobile robot selects the scene node on the data panel that has the largest probability mass associated with it, and then works its way down the data panel to extract the edges associated with that scene node. For example, for the expectation map of Fig. 5a and the image of Fig. 5b, all the data edges corresponding to the scene level node with the largest probability mass are shown in Fig. 5c. From the data edges thus extracted, the navigation module retains those that have labels with probability masses exceeding some high threshold, usually 0.9. These edges and their labels are then used for self-location.

The actual calculation of the robot's location is carried out by keeping track of two coordinate systems: the world coordinate system, represented by $W^3$, in which the hallways are modeled, and the robot coordinate system, represented by $R^3$, which translates and turns with those motions of the robot. The camera is calibrated in $R^3$; the calibration parameters make it possible to calculate the line of sight in $R^3$ to any pixel in the image. The problem of robot self-location is to compute the position and the orientation of $R^3$ with respect to $W^3$. In our work, we have assumed that the origin of $R^3$ always stays in the xy-plane of $W^3$ and that the z-axes of both coordinate systems are parallel and designate the vertical. By the orientation of $R^3$ we mean the angular rotation of the xy-plane of $R^3$ with respect to the xy-plane of $W^3$.

We have shown in [Lopez-Abadia and Kak 1989] how the edge correspondences, as returned by PSEIKI, between the camera image and the expectation map, can be used to solve the problem of self-location. To summarize the procedure described in [Lopez-Abadia and Kak 1989], we decompose the problem of robot self-location into two sub-problems: the problem of finding the orientation of $R^3$ and the problem of the finding the coordinates of the origin of $R^3$ in the xy-plane of $W^3$. As shown in [Lopez-Abadia and Kak 1989], the orientation of the robot can be found from a single edge in the camera image provided that edge has been identified as one of the horizontal edges by PSEIKI. If more than one horizontal edge is available, a weighted average is taken of the orientation results produced by the different edges, the weighting being a function of the belief values associated with the edges. A couple of different approaches are used simultaneously to compute the coordinates of the origin of $R^3$ in $W^3$. One of these approaches makes use of the fact that it is possible to compute the perpendicular distance of the origin of $R^3$ from a single edge in the image if that edge has been identified as a horizontal edge. Therefore, if PSEIKI can show matches for two non-parallel horizontal lines in the model, the world coordinates of the origin of $R^3$ are easily computed. The second approach is capable of using any two image edges, provided they correspond to two non-parallel lines in the model, for computing the location of the origin of $R^3$. Again, the results produced by both these approaches, for all possible pairs of edges satisfying the necessary conditions, are averaged using weights that depend upon the beliefs associated with the edges.

## 7. CONCLUDING REMARKS

A unique feature of PSEIKI is that it enforces model-generated constraints at different levels of abstractions, and it does so under a very flexible flow of control made possible by the blackboard implementation. Could we have used, say, a relaxation-based approach for the same end purpose, that of labeling the edges or regions in an image with labels from a model? We believe not, for the following reason: To parallel PSEIKI's

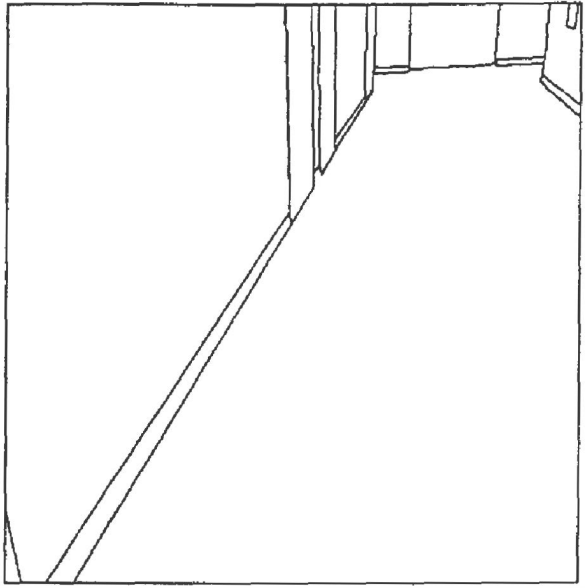
(a)

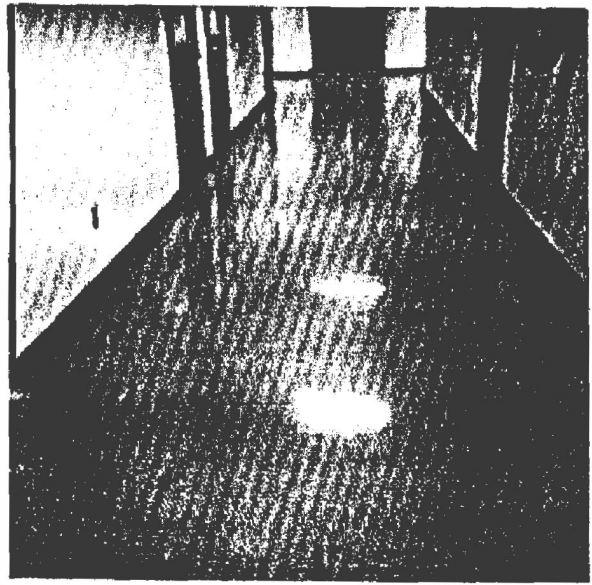
(b)

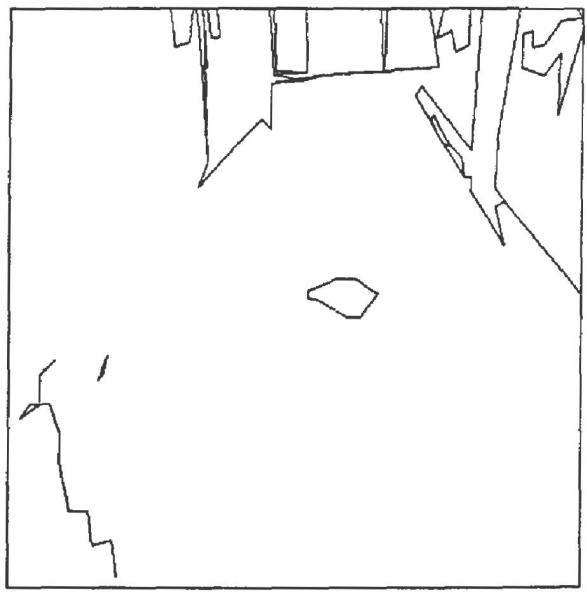
(c)

Fig. 5: While (a) shows the expectation map, (b) shows the actual camera image for an exercise in self-location. Mis-registration between the two is evident. Shown in (c) are the data edges from the image of (b) corresponding to the scene node with the largest probability mass associated with it.

competence at reasoning under uncertainty, one would have to use probabilistic relaxation. However, we do not believe it would be an easy feat to make the implementation of probabilistic relaxation as model-driven as is PSEIKI. In other words, it would be hard to incorporate in probabilistic relaxation as much model knowledge as we can in PSEIKI.

To give the reader an idea of the size of PSEIKI, the OPS83 code for the blackboard is about 10,000 lines long, with another 10,000 lines of C code for many functions called by the rules and for the image pre-processor. Currently, on a SUN3 workstation, it takes PSEIKI about 15 minutes to process one image, which, as was mentioned before, is one of the reasons why the mobile robot does not attempt self-location continuously, but only on a need basis -- the frequency of need depending on the quality of the odometry.